\documentclass{article}

\usepackage[final]{neurips_2019}
\usepackage{natbib}
\usepackage[T1]{fontenc}
\usepackage[utf8]{inputenc}

\usepackage{times}
\usepackage{epsfig}
\usepackage{graphicx}
\usepackage[tight,footnotesize]{subfigure}
\usepackage{amsmath}
\usepackage{amssymb}
\usepackage{booktabs} 
\usepackage{url}    

\usepackage{algorithmic}
\usepackage{algorithm}

\newtheorem{theorem}{Theorem}
\newtheorem{remark}{Remark}

\def\Z{\mathcal{Z}}

\def\N{\mathbb{N}}
\def\R{\mathbb{R}}

\def\V{\mathbb{V}}

\def\F{\mathcal{F}}
\def\G{\mathcal{G}}

\def\cost{\mathrm{cost}}

\def\D{\mathcal{D}}
\def\E{\mathcal{E}}

\def\mow{\mbox{\rm{MoW}}}

\usepackage{xcolor}

\begin{document}

\title{One-element Batch Training by Moving Window}

\author{
  Przemys\l{}aw Spurek\\
  \texttt{przemyslaw.spurek@uj.edu.pl} \\
  \And
  Szymon Knop\\
  \texttt{szymon.knop@doctoral.uj.edu.pl } \\
  \And
  Jacek Tabor\\
  \texttt{jacek.tabor@uj.edu.pl} \\
  \And
  Igor Podolak\\
  \texttt{igor.podolak@uj.edu.pl} \\
  \And
  Bartosz Wójcik\\
  \texttt{bartwojc@gmail.com} \\
}

\maketitle

\begin{abstract}
Several deep models, esp. the generative, compare the samples from two distributions (e.g. WAE like AutoEn\-co\-der models, set-processing deep networks, etc) in their cost functions. Using all these methods one cannot train the model directly taking small size (in extreme -- one element) batches, due to the fact that samples are to be compared.

We propose a generic approach to training such models using one-element mini-batches. The idea is based on splitting the batch in latent into parts: previous, i.e. historical, elements used for latent space distribution matching and the current ones, used both for latent distribution computation and the minimization process. Due to the smaller memory requirements, this allows to train networks on higher resolution images then in the classical approach. 
\end{abstract}

\section{Introduction}
In recent years a number of deep neural network models which use cloud of points/samples and invariant with respect to the permutation order function in training process was constructed. One distinctive class of such consists of Wasserstein  autoencoder  WAE~\citep{tolstikhin2017wasserstein} or Cramer-Wold autoencoder CWAE~\citep{tabor2018cramer} models.  Both use elegant geometric properties of the Wasserstein~\citep{arjovsky2017wasserstein} and the Maximum Mean Discrepancy MMD~\citep{tolstikhin2017wasserstein} distances.  The metrics are often used to measure the distance between two samples -- reprehension of data-set and sample from  prior distribution. 

Distances between samples might also be used in adversarial networks. Generative moment matching network GMMN~\citep{li2015generative} is a deep generative model differs from generative adversarial model GAN~\citep{goodfellow2014generative} by replacing the discriminator with a two-sample test based on MMD. MMD GAN~\citep{arbel2018gradient,binkowski2018demystifying} is a modification of GMMN and classical GAN by introducing adversarial kernel learning techniques, as the replacement of a fixed Gaussian kernel in the original GMMN. 

Invariancy to the permutation order which works on cloud of point/samples is also used in the case of processing sets~\citep{maziarka2018deep,zaheer2017deep} or 3D point cloud~\citep{qi2017pointnet,zamorski2019generative}. Each point is processed by a neural network and than a simple symmetric function is used to aggregate the information from all the points. 

Similar approach is used in sliced generative models~\citep{knop2018sliced}. In a Sliced Wasserstein Autoencoder (SWAE)~\citep{kolouri2018sliced} authors use sliced-Wasserstein distance one dimensional projections. The methodology is to take the mean of Wasserstein distances between one-dimensional projections of data-set. 

\begin{figure}
\label{fig:slice_win}
\begin{center}
\includegraphics[height=2.5cm]{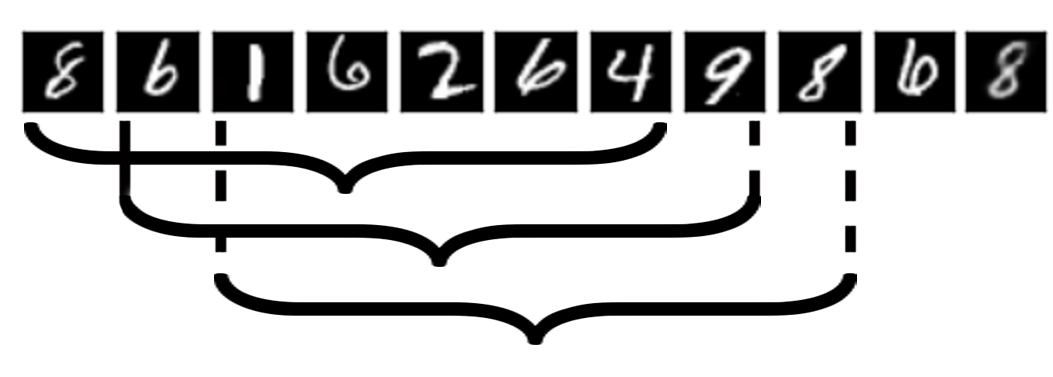}
\end{center}
  \caption{Batches constructed using moving windows.}
\label{fig:short}
\end{figure}

All above methods have an important limitation. That is, one cannot train the model by taking small (in extreme -- one element) batches, due to the fact that samples are to be compared. Contrarily, in case of adversarial training (e.g. GAN~\citep{goodfellow2014generative}, WGAN~\citep{arjovsky2017wasserstein}), flow-based generative models~\citep{dinh2014nice,kingma2018glow}) approaches training is essentially element-independent due to some additional function used: log-likelihood in flow models, discriminator in GAN, or the Lipschitz function in WGAN. In some sense, that function accumulates knowledge over previous batches.

At the same time, the impact of mini-batch size on generalization is being examined in depth. Experiments show that, for a classification problem, using small batch sizes give preferable training stability and generalization performance~\citep{masters2018revisiting}. In all experiments presented, the best results have been obtained with mini-batch sizes $m = 32$ or smaller, often as small as $2$ or $4$.

In this paper we show how one can train such models using one-element mini-batches. Each batch in latent is treated as a group of previous (historical) elements and the current mini-batch. The minimization process is applied only to the current elements group, while latent space normality (or any other prior set) is based on all. Solution is based on batches constructed using moving windows and in each new batch we have $k$ new elements and $n-k$ elements from previous batch, see Fig.~\ref{fig:slice_win}. Then, there is no need to propagate historical elements through the network as very close approximations of latent values corresponding to them are available from last iteration, provided careful training. The network is therefore trained using only $k$ elements, where it might even denote a single element. Consequently, our neural network is trained by using only $k$ elements (in extreme -- a single data element). 

In experiments using autoencoder generative models, we show that such learning process gives similar, often better, results to the classical approaches, while making it possible to train the network using large inputs, e.g.~large pictures.

Our contributions are the following:
\begin{enumerate}
\item we construct an optimizer \mow{}
which allows to use one-element batches during optimization in latent-models,
\item construct auto-encoder models using the defined MoW optimizer,
\item show experiments of proposed learning using high definition images. 
\end{enumerate}

\section{Related work}
Generative models introduced, both of variational and adversarially trained (e.g.~\citep{tolstikhin2017wasserstein,arjovsky2017wasserstein,goodfellow2014generative,kingma2014auto,makhzani2015adversarial}) have become very important, having proved to be highly highly usable in numerous fields, e.g.~\citep{gregor2015draw,heljakka2019towards,isola2017image,dosovitskiy2015learning,brock2016neural}. The maximum mean discrepancy learning MMD algorithm, ,which we approach here,was defined and dealt in depth, for several architectures, e.g. in~\citep{binkowski2018demystifying, dziugaite2015training,li2015generative}.

Work on high and super-fine resolutions require lots of memory to exploit a GPU efficiently, see e.g.~\citep{kingma2018glow,heljakka2018pioneer,larsen2015autoencoding,ledig2016photorealistic}. Our objective is to propose a new approach to the optimization process such that this memory requirements could be handled with. While big image approaches like in~\citep{heljakka2019towards} try to deal with large images by iterative multiplication of them learning from low to high resolutions, the amounts of memory needed are not handled with.  

The motivation of methodology in this paper is to propose an optimizer approach which could handle high memory requirements enabling high GPU bandwidth. The key is to take advantage of historical data computed. Several learning approaches use this paradigm. One is, naturally, the momentum optimization extending the regular Gradient Descent optimizer. 
Another is the Batch Normalization~\citep{ioffe2015batch}. Scale $\gamma$ and shift $\beta$ parameters are optimized using estimates of mini-batch means and variances computed during training. With tiny batch sizes, these estimates might become less accurate approximations of the true mean and variance used for testing~\citep{ioffe2017batch}. It has been shown to significantly improve training performance and has now become a~standard component of most state-of-the-art networks.

A specific example of one example per minibatch is that in Glow network~\citep{kingma2018glow}. Instead of batch normalization, authors use an activation normalisation, to perform an affine transformations of activations using scale and bias parameters per channel initialized such that the post-actnorm activations per channel have zero mean and unit variance given initial mini-batch being a data-dependent initialization~\citep{salimans2016weight}.

\section{Moving window in latent models}
We shall now present the moving window model in a general setting. Consider an $n$-element vector of elements $(q_i)_{i=1..n}\sim{}Q(D)$, where $Q$ can be thought of as a data generator producing a queue of examples from some data set $D$. We consider a cost function for $(q_i)$
\begin{equation} \label{eq:1}
\cost(\theta;q_1,\ldots,q_n)=\F(\E_\theta(q_1),\ldots,\E_\theta(q_n);V)+\sum_{i=1}^n \G_\theta(q_i),
\end{equation}
where $Z$ denotes the latent space, $\E:\R^N \to Z$ and $V$ is a sample vector generated by a given fixed random vector $\V$, while $\theta$ are weight parameters. Since the batch is sampled randomly, without loss of generality\footnote{Replacing $\F$ with $\tilde \F$ given by its mean over all permutations: $\tilde \F(z_1,\ldots,z_n;V)=\frac{1}{n!}\sum_{\sigma} \F(z_{\sigma(1)},\ldots,z_{\sigma(n)};V)$.} we may reduce to the case when $F$ is invariant with respect to permutations
\begin{equation*}
\F(z_{\sigma(1)},\ldots,z_{\sigma(n)};V)=\F(z_1,\ldots,z_n;V)
\end{equation*}
where $\sigma$ is an arbitrary permutation, while $z_i=\E(q_i)$. The above setting is correct, but to be effective with respect to memory savings, the dimension of the latent needs to be smaller then that of the input space.

\begin{remark}
One can observe, see next section, that typical autoencoder based generative models can be written in the above setting \eqref{eq:1}, where $\G$ is the reconstruction error and $\F$ is used to compare the sample constructed by the generator with the prior latent distribution (see the discussion in the next section). 

E.g, a Wasserstein GAN\citep{arjovsky2017wasserstein}, sampling examples $x$ and vector $V$, using a generator $\D$ to decode both and compare (the $\F$ function), to train both generator and discriminator with $\G$ cost defined with its rate of correct recognition, can be just described as above. Similarly an MMD~GAN~\citep{binkowski2018demystifying}.
\end{remark}

We are ready to proceed with the introduction of the $\mow_k(n)$ optimization procedure, where $n$ as before denotes the theoretical mini-batch size, as in the standard procedure, and $k$ is the real batch size, i.e. the number of examples read at each step and kept in the memory. Thus $\mow_1(n)$ will be using in practice one-element batches. By $\eta$ we denote the learning rate.

Now let $X=(x_i)_{i=1,\ldots,\infty}$ denote a fixed infinite random sample generated from the data distribution (in practice it comes by drawing with repetitions from the original data-set). In that sense we can identify $X$ with the dataset generator $Q(D)$ output. We also choose a random sequence $(V_i)$ generated from some given prior $P_\Z$.

Our objective is to define inductively the sequence $(\theta_l)$ of weight parameter vectors and of ``historical'' latent elements $(z_l)\in{}Z$.

STEP 1 (initialization). As is common, we start with some (typically randomly chosen) initial weights $\theta_0$. Given $\theta_0$, we define the first $(n-k)$ elements of the sequence $(z_j)$ using formula 
\begin{equation*}
z_j=\E_{\theta_0}(x_j) \mbox{ for }j=1,\ldots,(n-k).
\end{equation*}

STEP 2 (recursive step $l \to l+1$). Suppose that we have already defined $\theta_j$ for $j=0,\ldots,l$ and $z_j$ for $j=1,\ldots,(n-k+lk)$, i.e. $k$ elements for each of the past $l$ iterations. Put $m=n-k+lk$. We consider now the modification of the cost function \eqref{eq:1} so that all gradients on elements $(x_j)$ for $j=1,\ldots,m$ are frozen. Namely, we put 
\begin{equation*}
\cost_{l+1}(\theta)
=\F_{\theta}(\underbrace{z_{m-n+k+1},\ldots,z_m}_{n-k}, \underbrace{\E_{\theta}(x_{m+1}),\ldots,\E_{\theta}(x_{m+k})}_{k}; V_{l+1})+\sum \limits_{j=m+1}^{m+k}\G_{\theta}(x_j).
\end{equation*}
Notice that we have dropped from the original cost function the first $(n-k)$ elements of the sum $\sum_{j=m-n+k+1}^{m+k}\G_{\theta}(x_j)$ as the reconstruction error shall not be minimized for the historical elements now. We minimize the above with an arbitrary standard gradient descent procedure starting at $\theta_{l}$, e.g.:
\begin{equation*}
\theta_{l+1}=\theta_{l}+\eta\nabla \cost_{l+1}(\theta_{l}).
\end{equation*}
We update the sequence $(z_j)$ by putting
\begin{equation*}
z_j=\E_{\theta_{l+1}}(x_j) \mbox{ for }j=(m+1),\ldots,(m+k).
\end{equation*}

    \begin{algorithm}[H]
    \caption{\mow{} optimization} 
    \label{alg2} 
    \begin{algorithmic} 
    \STATE {\bf Require: }
    \STATE\hspace\algorithmicindent Initialize $l=0$
    \STATE\hspace\algorithmicindent Initialize $\eta$
    \STATE\hspace\algorithmicindent Initialize the parameters $\theta_l$
    \STATE\hspace\algorithmicindent Sample batch  $( q_1, \ldots, q_{n-k}) \sim Q $ 
    \STATE {\bf Initialization: }    
    \STATE\hspace\algorithmicindent Calculate 
    $
    z_j=\E_{\theta_0}(q_j) \mbox{ for }j=1,\ldots,(n-k)
    $

    \WHILE{ not converged }
    \STATE\hspace\algorithmicindent Sample $( q_{n-k+1}, \ldots, q_n )$ from the training set
    \STATE\hspace\algorithmicindent Sample $ V $ from the random variable $\V$
    \STATE\hspace\algorithmicindent Define
    $$
    \cost_{l+1}(\theta)= 
    \E_{\theta}(q_{n-k+1}),\ldots,\E_{\theta}(q_{n}); V) 
    +\sum \limits_{j=n-k+1}^{n}\G_{\theta}(q_j)
    $$
    \STATE\hspace\algorithmicindent
    Update $\theta$ by: 
    $
    \theta_{l+1}=\theta_{l}+\eta\nabla \cost_{l+1}(\theta_{l})
    $
    \STATE\hspace\algorithmicindent
    Update $l = l+1$    
    \STATE\hspace\algorithmicindent
    Update 
    $
    (z_1, \ldots, z_{n-k}) = ( z_{k+1}, \ldots, z_n, \E_{\theta_l}(q_{n-k+1}), \ldots,  \E_{\theta_l}(q_{n}))
    $        
    \ENDWHILE
    \end{algorithmic}
    \end{algorithm}

The \mow{} pseudo-code is given in Algorithm~\ref{alg2}.

In the following theorem we show that the \mow{} procedure gives the correct approximation of the gradient descent method.

\begin{theorem}
Let $C:\theta \to \R$ denote the expected value of the cost function, where $\theta$ denote the weight space, i.e.:
$$
C(\theta)=\mathbb{E}[ \cost(\theta;q_1,\ldots,q_n,V)| (q_i) \sim Q(D),V \sim \V].
$$
Let $S:[0,T] \to \theta$ denote the exact solution of the gradient optimization process starting from $\theta_0$, $S(0)=\theta_0$:
\begin{equation} \label{eq:2}
S'(t)=-\frac{k}{n}\nabla C(S(t)).
\end{equation}

Let $S_\eta:[0,T]_\eta \to \theta$ denote the solution given by $\mow(k,n)$ with step size $\eta$ with 
\begin{itemize}
\item fixed random choice of sequence $(x_j)$ from the data set
generated by $Q(D)$,
\item random sample $(V_j)$ from the random variable,
\item initial starting weight $\theta_0$: $S_\eta(0)=\theta_0$, 
\end{itemize}
where $[0,T]_\eta:=\{k\eta \, | \, k \in \N,k\eta \leq T\}$ is the discretization of time (with respect to step size $\eta$).
Then 
\begin{equation} \label{eq:25}
\lim_{\eta \to 0} S_\eta=S.
\end{equation}
\end{theorem}

{\em Proof. } We first extend $S_\eta$ in an affine way to the whole interval $[0,T]$. Now, for an arbitrary $t \in [0,T)$ we are going to show that 
\begin{equation} \label{eq:3}
\frac{S_\eta(t+h)-S_\eta(t)}{h} =-\frac{k}{n} \nabla C(S_\eta(t))+o(h),
\end{equation}
for sufficiently small $\eta$.
If \eqref{eq:3} is valid, $S_\eta$ (for $\eta$ sufficiently close to zero) is an approximate solution to the discretization \eqref{eq:2}.
Since $S_\eta(0)=\theta_0$, by the uniqueness of the solutions of differential equations, we obtain \eqref{eq:25}, and consequently the assertion of the theorem.

Let us now proceed to the proof of \eqref{eq:3}. We take $h$ small enough so that the changes in the weights are minimal, i.e. $S_\eta(s)\approx S_\eta(t)$ for $s \in [T,T+h]$. Let us consider the procedure defining $S_\eta$ described in the first part of the section, where we assume that $\eta \ll h$, i.e. we assume that $\eta=K h$ for some large $K$. We investigate the iterative process defining $S_\eta$. Let us first observe, that (we apply the notation used in the introduction to the model):
\begin{equation*}
\begin{array}{l}
\frac{\partial}{\partial \theta} \cost_{l+1}(\theta)
= \sum \limits_{j=n-k+1}^n \frac{\partial \F_\theta}{\partial r_j}(z_{m-n+k+1},\ldots,z_m,\E_\theta(x_{m+1}),\ldots,\E_\theta(x_{m+k});V_{l+1}) \frac{\partial \E_\theta}{\partial \theta}(x_j)\\
[2ex]
\hskip6.5em+\sum \limits_{j=m+1}^{m+k}\frac{\partial G_\theta}{\partial \theta}(x_j).
\end{array}
\end{equation*}
Since $\theta$ does not, almost, change during the optimization process for the time in the interval $[t,t+h]$, $\E_\theta x_j  \approx z_j$, and therefore we can approximate the above by
\begin{equation} \label{eq:27}
\begin{array}{l}
d_n=\sum \limits_{j=n-k+1}^n \frac{\partial \F_\theta}{\partial r}(\E_\theta x_{m-n+k+1},\ldots,\E_\theta x_{m+k};V_{l+1}) \frac{\partial \E_\theta}{\partial \theta}(x_j) +\sum \limits_{j=m+1}^{m+k}\frac{\partial G_\theta}{\partial \theta}(x_j),
\end{array}
\end{equation}
where we use the notation $\frac{\partial F_\theta}{\partial r}$ to denote an arbitrary $\frac{\partial F_\theta}{\partial r_j}$ (they are all equal by the assumptions). To obtain approximation of the mean derivative over the interval $[t,t+h]$ we take the mean  
$\frac{1}{K}\sum \nolimits_{i=n}^{n+N-1} d_n$.

Now the derivative of the cost function $C$ is given by 
\begin{equation*}
C'(\theta)=\mathbb{E}\left[ \sum_{i=1}^n \frac{\partial \F_\theta}{\partial r}(\E_\theta q_1,\ldots,\E_\theta q_n;V)\frac{\partial \E_\theta}{\partial \theta}+\sum_{i=1}^n \frac{\partial G_\theta}{\partial \theta}(x_j) | (q_i) \sim Q(D),V \sim \V\right].
\end{equation*}
One can easily observe, that since in \eqref{eq:27} every component has only $k$ factors, while in the above formula every sum has $n$ factors, taking $K$ large we obtain \eqref{eq:3}. 
\hfill$\Box$

\section{Experiments} \label{se:ex}
In this section we empirically validate the proposed training based on moving windows\footnote{The code is available  \url{https://github.com/gmum/MoW}}. We use three datasets: CELEB~A, CIFAR-10,  and MNIST (for the Fashion MNIST experiments see~B). We show the properties of our training strategy in the case of three methods: CWAE~\citep{tabor2018cramer}, WAE-MMD~\citep{tolstikhin2017wasserstein} and SWAE~\citep{kolouri2018sliced}.  As we shall see, it is possible to train such models, by using batches containing only one element.

For convenience of the reader we start form short description of CWAE~\citep{tabor2018cramer}, WAE-MMD~\citep{tolstikhin2017wasserstein} and SWAE~\citep{kolouri2018sliced}.
Let $X=(x_i)_{i=1..n} \subset \R^N$ be a given data set, which can be considered as sample from true, though unknown, data distribution $P_X$. The basic aim of an autoencoder is to transport the data to a (typically, but not necessarily) less dimensional latent space $\Z=\R^D$ with reconstruction error as small as possible. Thus, we search for an encoder $\E\colon\R^N \to \Z$ and decoder $\D\colon\Z \to \R^N$ functions, which minimize some reconstruction error, e.g.
$mse(X;\E,\D)=\frac{1}{n}\sum_{i=1}^n \|x_i-\D(\E x_i)\|^2$.

An autoencoder based generative model extends AE by introducing a cost function that makes the model generative, i.e. ensures that the data transported to the latent space $\Z$ conforms to some given (frequently  Gaussian) prior distribution $P_{\Z}$.  A usual way to ensure it is through adding to $mse(X;\E,\D)$ a regularization (using appropriate $\lambda>0$) term that penalizes dissimilarity between the distribution of the encoded data $P_{\E(X)}$ and the prior $P_\Z$.

WAE~\citep{tolstikhin2017wasserstein} is a classical autoencoder model which uses the Wasserstein metric to measure the distance between two samples -- latent representation of dataset and sample from  prior distribution:
\begin{equation*}
cost(X; \E, \D) =  mse(X;\E, \D) + \lambda\cdot d_{WAE}(\E(X), Z),
\end{equation*}
where $Z$ is a sample form the prior distribution $P_\Z$.

Another autoencoder based model which uses a similar approach is the CWAE~\citep{tabor2018cramer}. In CWAE authors use Cramer-Wold distance
between latent representation and prior distribution $N(0,I)$:
\begin{equation*}
cost(X; \E, \D) = mse(X;\E, \D) +\lambda\cdot d_{CWAE}(\E(X), N(0,I)).
\end{equation*}

SWAE~\citep{kolouri2018sliced} is a modification of WAE relying  on the use of sliced Wasserstein  distance. It takes the mean of the Wasserstein distances between one-dimensional projections of $\E(X)$ and $Z$ (sample form prior $P_\Z$, usually $N(0,I)$). Note that SWAE, similarly to WAE, also needs sampling from $P_\Z$. Consequently in SWAE two types of sampling are applied: sampling over one-dimensional projections and sampling from the prior distribution
\begin{equation*}
cost(X; \E, \D) =  mse(X;\E, \D) + 
\lambda\cdot \frac{1}{k} \sum\nolimits_{i=1}^k d_{SWAE}( v_i^T \E(X), N(0,1)),
\end{equation*}
for $k$ one-dimensional projections on the spaces spanned by the unit vectors $v_i \in \R^D$ for $i= 1,\ldots,k$ and one-dimensional Wasserstein distance $d_{SWAE}$.

In the experiment we use two basic architecture types. Experiments on MNIST use a feed-forward network for both encoder and decoder, and a 20 neuron latent layer, all using ReLU activations. For CIFAR-10, and CELEB~A data sets we use convolution-deconvolution architectures. Please refer to Supplementary materials, Section~A for full details. Additional experiments on FASHION MNIST set are available in Supplementary materials, Section~C.

In our experiments we use classical Gradient descent minimization. This is because advanced models, like Adam, in itself use history to evaluate next steps. Thus, the results would not be clear to separate the results of the MoW and the basic optimizer. On the other hand,  results using Adam optimizer are presented in Supplementary materials, Section~B.

We evaluated three types of batches, all with $n=64$ overall elements. First, we use classical batches~$k=n$ (classical method). Second, we use batches containing only one new element $k=1$. The third model uses $k=32$.

\begin{figure*}[htb]
\centering
\begin{tabular}{@{}c@{}c@{}c@{}c@{}}
 &  Test interpolation &  Test reconstruction &  Random sample \\
\rotatebox{90}{ \ WAE-MMD} & \,
\includegraphics[height=3.5cm]{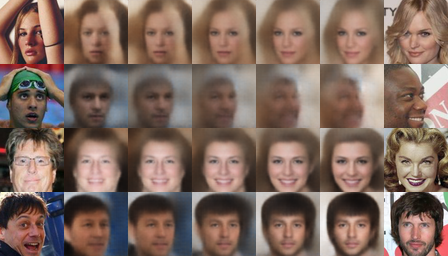} & \,
\includegraphics[height=3.5cm]{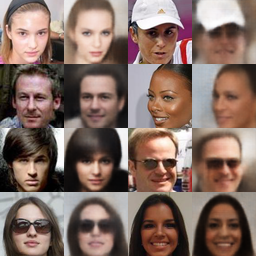}& \, 
\includegraphics[height=3.5cm]{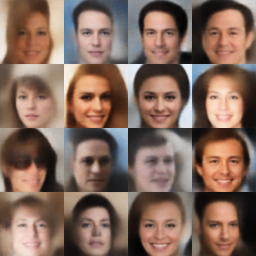} \\ 
\rotatebox{90}{ \ WAE-MMD $k=1$} & \,
\includegraphics[height=3.5cm]{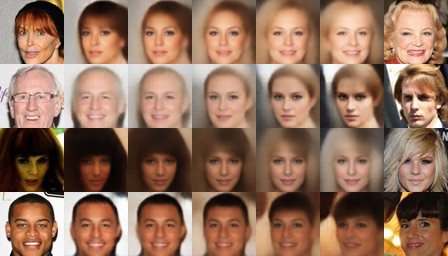} & \,
\includegraphics[height=3.5cm]{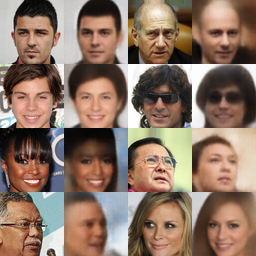}& \, 
\includegraphics[height=3.5cm]{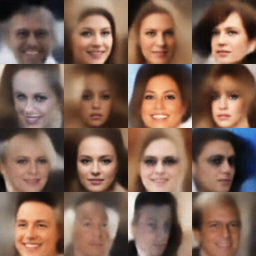}
\end{tabular}
\caption{Comparison between the MoW one-element mini-batch training procedure ($n=64$, $k=1$) with standard ($n=k=64$) on the CELEB~A data. \textbf{Left:} Interpolations between two examples from the test distribution (left to right, in each row). The first and the last images in each row are examples from the dataset. \textbf{Middle:} Reconstruction of examples from the test distribution; odd rows correspond to the real test points. \textbf{Right:} Reconstructed examples from a random samples from the prior distribution.}
\label{fig:celeb}
\end{figure*}

The quality of a generative model is typically assessed by examining generated samples and by interpolating between samples in the hidden space. We present such a~comparison between our strategies in WAE-MMD architecture in Fig.~\ref{fig:celeb}. 

For each model we consider: interpolation between two random examples from the test set (leftmost in Fig.~\ref{fig:celeb}), reconstruction of a~random example from the test set (middle column in Fig.~\ref{fig:celeb}), and a~sample reconstructed from a~random point sampled from the prior distribution (right column in Fig.~\ref{fig:celeb}). 
The experiment shows that there are no perceptual differences between batches containing one element ($n=64$, $k=1$) and classical approach ($n=k=64$).

\begin{table*}[htb]
\normalsize
\caption{Comparison between the moving window mini-batch training procedure in two versions ($n=64$, $k=1$ and $n=64$, $k=32$) with classical one ($n=k=64$). 
We used grid search over learning rate parameter for all models choosing optimal model in respect to WAE cost, i.e. sum of reconstruction error and the logarithm of  WAE distance. For CELEB~A, see FID scores in Fig.~\ref{fig:wae_celeba_comparison}. }
\begin{center}
{\small
\begin{tabular}[width=\textwidth]{lllrrrrr}  
\toprule
Data set & Method & Learning & CWAE & WAE & Rec. & WAE & FID \\
 &  & rate & distance & distance & error & cost & score\\
\midrule             
MNIST  & WAE $n=64$, $k=64$ & 0.01 & 0.026 & 0.009 & 5.495 & 5.503 & 48.491 \\ 
         & WAE $n=64$, $k=32$ & 0.005 & 0.027 & 0.009 & 5.426 & 5.436 & 49.218 \\
         & WAE $n=64$, $k=1$ & 0.0025 & 0.154 & 0.306 & 4.894 & 5.200 & 54.753 \\
\midrule             
CIFAR10    & WAE $n=64$, $k=64$ & 0.0025 & 0.019 & 0.023 & 25.248 & 25.271 & 189.711 \\
         & WAE $n=64$, $k=32$ & 0.005 & 0.017 & 0.019 & 24.831 &  24.851 &  156.142 \\
         & WAE $n=64$, $k=1$ & 0.001 & 0.083 & 0.192 & 26.177 & 26.369 & 200.684 \\
\midrule             
CELEB~A & WAE  $n=64$, $k=64$ & 0.005 & 0.016 & 0.008 & 116.873 & 116.881 & 62.234 \\
         & WAE $n=64$, $k=32$ & 0.005 & 0.017 & 0.009 & 117.256 & 117.266 &  60.605 \\
         & WAE $n=64$, $k=1$ & 0.001 & 0.064 & 0.124 & 116.306 &  116.431 & 74.880 \\
\bottomrule
\end{tabular}
}
\end{center}
\label{tab:comp_1}
\end{table*}

\begin{figure*}[htb]
    \centering
    \includegraphics[width=1\textwidth]{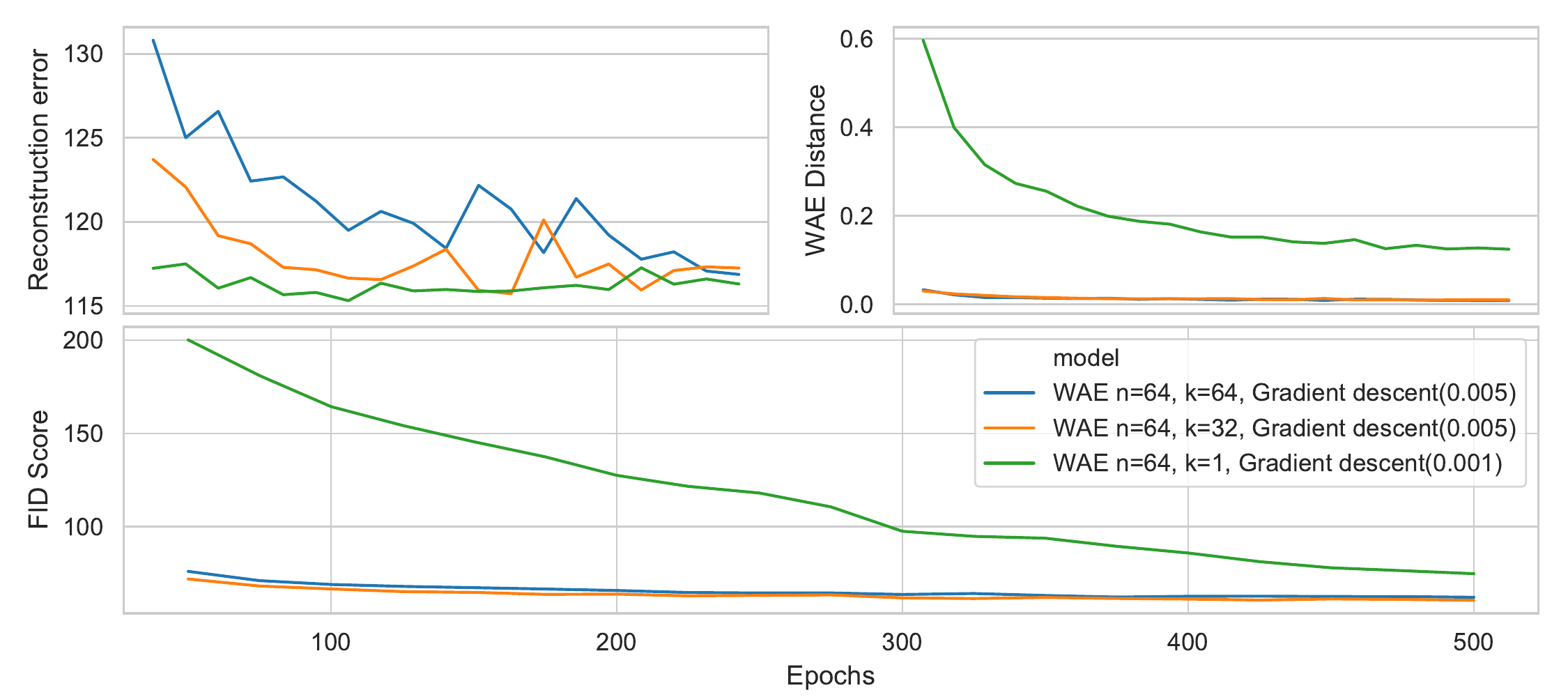}
    \caption{Comparison between the moving window mini-batch training procedure in two versions  ($n=64$, $k=1$ and $n=64$, $k=32$) with classical one ($n=k=64$) (all WAE and CELEB~A). }
    \label{fig:wae_celeba_comparison}
\end{figure*}

In order to quantitatively compare between our moving window mini-batch training procedure with the classical one, we use the Fr\'{e}chet Inception Distance (FID)~\citep{heusel2017gans}. WAE-MMD, CWAE and SWAE methods are used. Thus, in addition to FID, we will also report the reconstruction error and WAE distances. Results for WAE are presented in Fig.~\ref{fig:wae_celeba_comparison} and Tab.~\ref{tab:comp_1}. The results for CWAE and SWAE w presented in Supplementary materials, Section C. 

In Fig.~\ref{fig:wae_celeba_comparison} we report for CELEB~A data set: the FID score, reconstruction error and WAE distances during learning process. No essential differences between batches containing one element ($n=64$, $k=1$) and classical approach ($n = 64$, $k=0$) can be seen. Moreover, adding information from previous batches allows us to obtain better FID score on CIFAR~10 dataset. The method using only one new element in batches gives comparable results in grid search procedure. 

\begin{figure*}[htb]
\centering
\begin{tabular}{@{}c@{}c@{}c@{}c@{}}
 &  Test interpolation &  Test reconstruction &  Random sample \\
\rotatebox{90}{ \ CWAE $k=1$} & \,
\includegraphics[height=3.5cm]{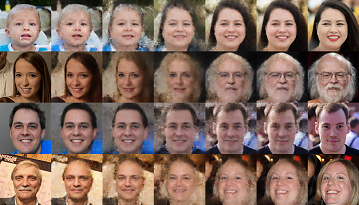} & \,
\includegraphics[height=3.5cm]{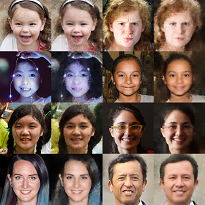}& \, 
\includegraphics[height=3.5cm]{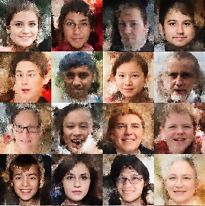} \\ 
\end{tabular}
\caption{Result of the moving window one-element mini-batch MoW training procedure ($n=64$, $k=1$)  on Flickr-Faces-HQ (FFHQ) database ($1024\times 1024$ pixels). \textbf{Left:} Interpolations between two examples from the test set (in rows). The first and the last images in each row are examples from the dataset. \textbf{Middle:} Reconstruction of examples from the test set; odd columns correspond to the real test points. \textbf{Right:} Random samples from the prior latent distribution.}
\label{fig:celeb_max}
\end{figure*}

Next, we show that our approach allows us to train our model on high resolution images from Flickr-Faces-HQ (FFHQ) database. FFHQ consists of 70,000 high-quality PNG images at $1024 \times 1024$ resolution and contains considerable variation in terms of age, ethnicity and image background.

The quality of our generative model is presented in Fig.~\ref{fig:celeb_max}: interpolation between two random examples from the test set, reconstruction of a~random example from the test set, and a~sample reconstructed from a~random point sampled from the prior latent distribution.

The experiment shows that it is possible to train our model on high resolution images directly. Therefore, we do not need any modification of the architecture like in progressively growing strategy~\citep{heljakka2018pioneer,heljakka2019towards} where the neural network architecture is increased in each iteration. In the case of Pioneer network authors trained the network progressively through each intermediate
resolution until reach the target resolution ($64 \times 64$, $128 \times 128$, or $256 \times 256$) on classical CelebA-HQ. We show that our approach can process high-resolution images directly.

\section{Conclusions} 
Training generative networks for high-dimensional problems, e.g. images, can be cumbersome. The dimensionality has a particular impact on the GPU memory used and a given card may not be able to store more than one example. On the other hand, the training methods which compare the latent distribution $P_{\E(X)}$ with some prior $P_{\Z}$ require that several training examples are used in a mini-batch. The examples occupy a lot of memory.

On the other hand, the latent dimension $D$ is usually, except for some very specific models, much smaller than the input example size. Therefore we proposed to use a buffer of some recent latent activation vectors together with a input single example. Past activations together with the current one would be used for minimizing given distribution distance metric $d(\E(X), P_{\Z})$, while the current example would drive the reconstruction error down.

We have proposed the \mow{} optimizer, that can work with as little as single input example. We have shown that such procedure approximate classical mini-batch strategy. To show that the proposed approach is useful in practice, we have performed several experiments, together with those on high-dimensional FFHQ images. The results are comparable to those while training on bigger mini-batches.

\bibliography{egbib}

\newpage

{\center \Large

Supplementary materials \\
One-element Batch Training by Moving Window

}

\appendix
\section{Architecture details}\label{app:architectures}
The following feedforward (MNIST) and convolution-deconvolution autoencoder architectures were used:

MNIST:
\begin{itemize}
\item[] \textbf{input} $\mathbb{R}^{28\times28}$
\item[] \textbf{encoder} 
\begin{itemize}
\item[] $3\times$ fully connected ReLU layers, 200 neurons each. 
\item[] $1\times$ fully connected layer with 20 neurons.
\end{itemize}
\item[]\textbf{latent} 20-dimensional
\item[]\textbf{decoder} 
\begin{itemize}
\item[] $2\times$ fully connected ReLU layers, 200 neurons each. 
\item[] $1\times$ fully connected layer with 200 neurons and sigmoid activation.
\end{itemize}
\end{itemize}

CelebA:
\begin{itemize}
\item[] \textbf{input} $\mathbb{R}^{64\times64\times3}$
\item[]\textbf{encoder}
\begin{itemize}
\item[] $4\times$ convolution layers with $4\times4$ filters and $2\times2$ strides (consecutively 32, 32, 64, and 64 output channels), all ReLU activations,
\item[] $2\times$ fully connected layers (1024 and 256 ReLU neurons)
\item[] $1\times$ fully connected layer with 32 neurons.
\end{itemize}
\item[]\textbf{latent} 32-dimensional
\item[]\textbf{decoder}
\begin{itemize}
\item[] $2\times$ fully connected layers (256 and 1024 ReLU neurons), 
\item[] $3\times$ transposed-convolution layers with $4\times4$ filters with $2\times2$ strides (consecutively 64, 32, 32 channels) with ReLU activation,
\item[] $1\times$ transposed-convolution $4\times4$ with $2\times2$ stride, 3 channels, and sigmoid activation.
\end{itemize}
\end{itemize}

CIFAR-10
\begin{itemize}
\item[]\textbf{input} $\mathbb{R}^{32\times32\times3}$
\item[]\textbf{encoder}
\begin{itemize}
\item[] $4\times$ convolution layers with $2\times2$ filters, the second one with $2\times2$ strides, other non-strided (3, 32, 32, and 32 channels) with ReLU activation,
\item[] $1\times$ fully connected ReLU layer with 128 neurons,
\item[] $1\times$ fully connected layer with 64 neurons.
\end{itemize}
\item[]\textbf{latent} 64-dimensional
\item[]\textbf{decoder}
\begin{itemize}
\item[] $2\times$ fully connected $ReLU$ layers with $128$ and $8192$ neurons,
\item[] $2\times$ transposed-convolution layers with $2\times2$ filters (32 and 32 channels) and ReLU activation,
\item[] $1\times$ a transposed convolution layer with $3\times3$ filter and $2\times2$ strides (32 channels) and ReLU activation,
\item[] $1\times$ a transposed convolution layer with $2\times2$ filter (3 channels) and sigmoid activation.
\end{itemize}
\end{itemize}

Fashion MNIST:
\begin{itemize}
\item[] \textbf{input} $\mathbb{R}^{28\times28}$
\item[]\textbf{encoder}
\begin{itemize}
\item[] $4\times$ convolution layers with $4\times4$ filters, $2\times2$ strides and consecutively 128, 256, 512, and 1024 channels followed by ReLU activation,
\item[] $1\times$ a fully connected layer with 8 neurons.
\end{itemize}
\item[]\textbf{latent} 8-dimensional
\item[]\textbf{decoder} 
\begin{itemize}
\item[] $1\times$ a fully connected ReLU layer with $7\times7\times1024$ neurons,
\item[] $2\times$ transposed-convolution layers  with $4\times4$ filters, $2\times2$ strides and consecutively 512 and 256 channels,
\item[] $1\times$ a transposed-convolution layer with $4\times4$ filter and 1 channel and $tanh$ activation.
\end{itemize}
\end{itemize}

FFHQ:
\begin{itemize}
\item[] \textbf{input} $\mathbb{R}^{1024\times1024\times3}$
\item[]\textbf{encoder}
\begin{itemize}
\item[] $4\times$ convolution layers with $5\times5$ filters, $4\times4$ strides and consecutively 128, 256, 512, and 1024 channels followed by ReLU activation,
\item[] $1\times$ a fully connected layer with 64 neurons.
\end{itemize}
\item[]\textbf{latent} 64-dimensional
\item[]\textbf{decoder} 
\begin{itemize}
\item[] $1\times$ a fully connected ReLU layer with $16\times16\times1024$ neurons,
\item[] $3\times$ transposed-convolution layers  with $5\times5$ filters, $4\times4$ strides and consecutively 512, 256, and 128 channels,
\item[] $1\times$ a transposed-convolution layer with $5\times5$ filter and 3 channel with $tanh$ activation.
\end{itemize}
\end{itemize}


\section{Experiments using ADAM optimizer}\label{app:adam}

In this section we empirically validate the proposed training based on moving windows. We use Fashion MNIST dataset. We show the properties of our training strategy for three methods: CWAE~\citep{tabor2018cramer}, WAE-MMD~\citep{tolstikhin2017wasserstein} and SWAE~\citep{kolouri2018sliced}.  It turns out that it is possible to train these models using single-element batches.

Convolution-deconvolution architectures (see Supplementary materials, Section~A) together with Adam optimizer were used. We also used grid search for a learning rate parameter. We chose the optimal model in respect to WAE (and respectively CWAE and SWAE) cost, which consist of the sum of reconstruction error and the logarithm of WAE (and respectively CWAE and SWAE) distance.

\begin{figure}[htb]
\centering
\subfigure[WAE]
{\includegraphics[width=0.85\textwidth]{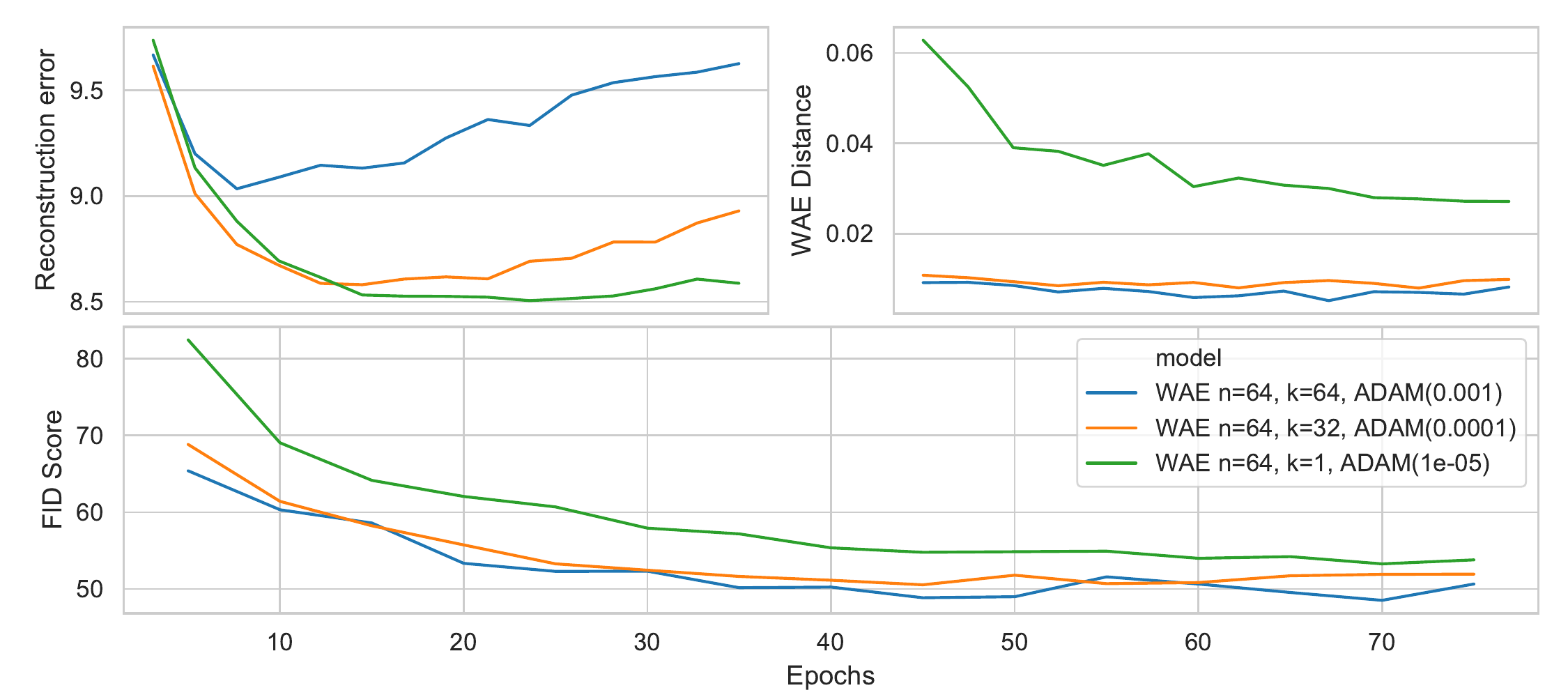}}
\subfigure[CWAE]
{\includegraphics[width=0.85\textwidth]{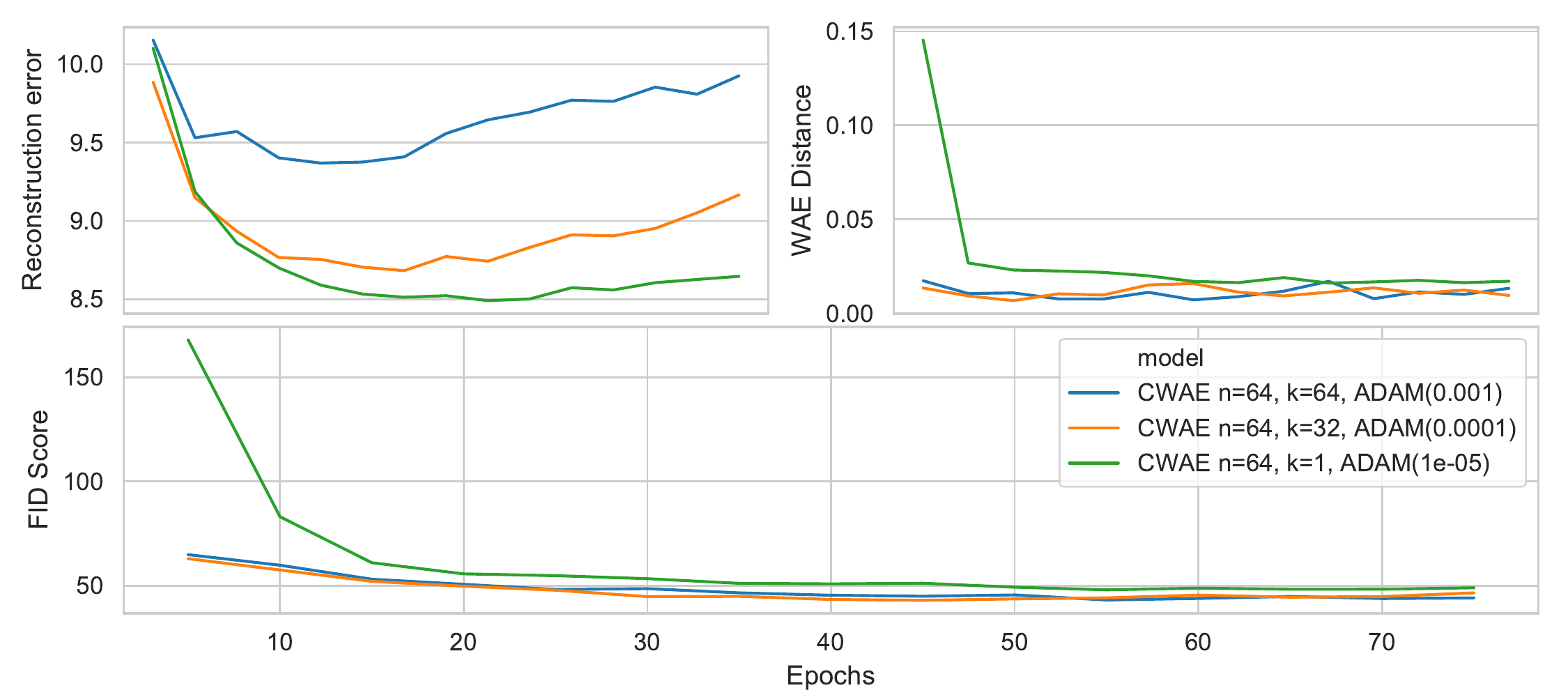}}
\subfigure[SWAE]
{\includegraphics[width=0.85\textwidth]{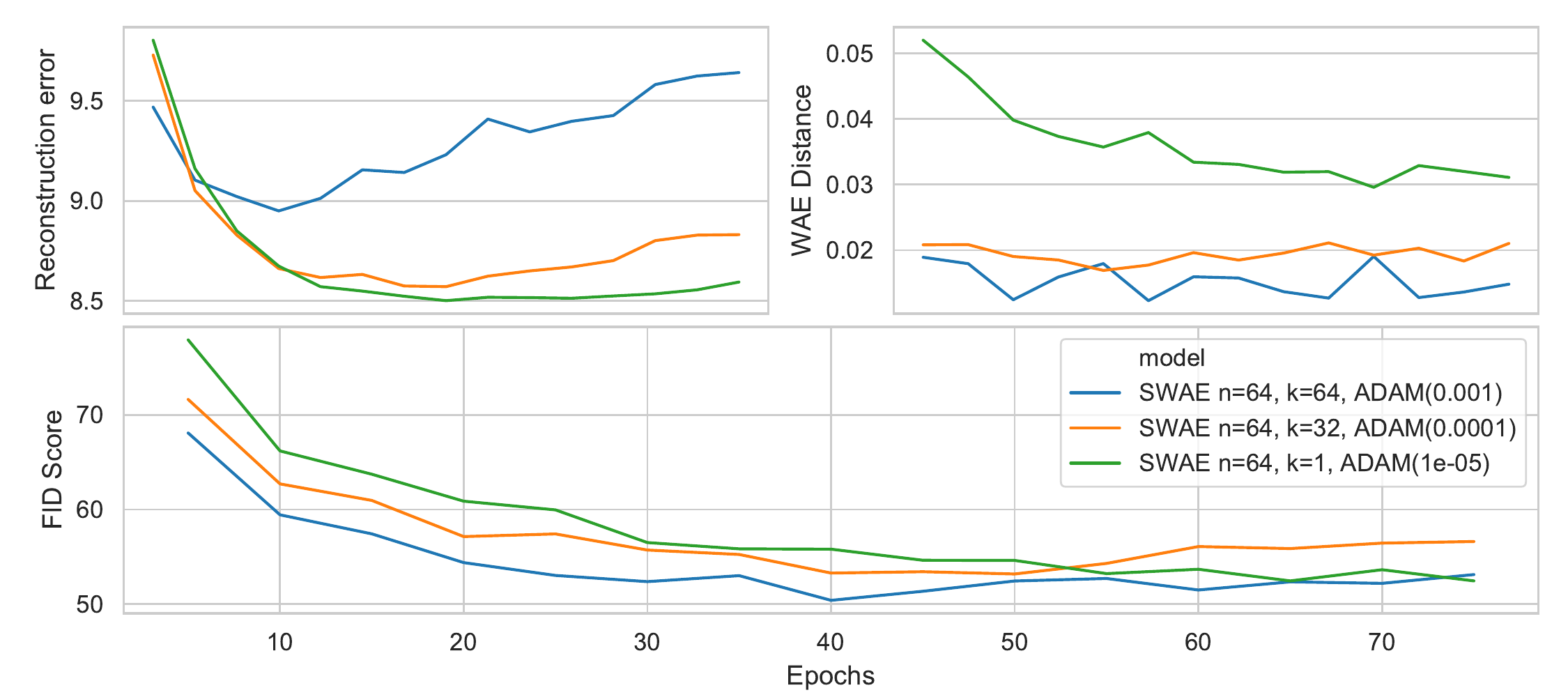}}    
\caption{Comparison between the moving window mini-batch training procedure in two versions  ($n=64$, $k=1$ and $n=64$, $k=32$) with classical one ($n=k=64$) while using ADAM optimizer on Fasion  MNIST dataset. We plot values of reconstruction error, WAE distance and FID score.
}
\label{fig:cwae_celeba_comparison}
\end{figure}

We evaluated three types of batches, see Fig.~\ref{fig:cwae_celeba_comparison} and Tab.~\ref{tab:comp_ADAM}.

\begin{table*}[htb]
\normalsize
\caption{Fashion MNIST, Adam - Comparison between the moving window mini-batch training procedure in two versions ($n=64$, $k=1$ and $n=64$, $k=32$) with classical one ($n=k=64$). }
\begin{center}
{\small
\begin{tabular}[width=\textwidth]{lllrrrrr}  
\toprule
Method & Learing & CWAE & WAE & Reconstruction & WAE & FID \\
& rate & distance & distance & error & cost & score\\
\midrule             
CWAE $n=64$, $k=64$ & 0.001 & 0.059 & 0.013 & 9.925 & 5.607 & 44.112 \\ 
CWAE $n=64$, $k=32$ & 0.0001 & 0.055 & 0.009 & 9.166 & 4.511 & 46.557 \\
CWAE $n=64$, $k=1$ & 0.00001 & 0.067 & 0.017 & 8.646 & 4.576 & 48.993 \\
\midrule             
WAE $n=64$, $k=64$ & 0.001 & 0.057 & 0.008 & 9.627 & 4.822 & 50.656 \\
WAE $n=64$, $k=32$ & 0.0001 & 0.058 & 0.009 & 8.931 & 4.312 & 51.940 \\
WAE $n=64$, $k=1$ & 0.00001 & 0.072 & 0.027 & 8.587 & 4.980 & 53.799 \\
\midrule             
SWAE  $n=64$, $k=64$ & 0.001 & 0.065 & 0.014 & 9.641 & 5.429 & 53.122 \\
SWAE $n=64$, $k=32$ & 0.0001 & 0.071 & 0.022 & 8.831 & 4.969 & 56.618 \\
SWAE $n=64$, $k=1$ & 0.00001 & 0.078 & 0.031	 & 8.594 & 5.123 & 52.453  \\
\bottomrule
\end{tabular}
}
\end{center}
\label{tab:comp_ADAM}
\end{table*}

\section{Additional experiments on CWAE and SWAE architecture}\label{app:ca_S_WAE}

In this section we empirically validate the proposed MoW training based on moving windows. We use three datasets: CELEB~A, CIFAR-10, MNIST. We show the properties of our training strategy in the case of CWAE~\citep{tabor2018cramer} and SWAE~\citep{kolouri2018sliced}.  

Two basic architecture types are used. Experiments on MNIST use a feed-forward network for both encoder and decoder, and a 20 neuron latent layer, all using ReLU activations. For CIFAR-10, and CELEB~A data sets we use convolution-deconvolution architectures (see Supplementary materials, Section~A).

In our experiments we use classical Gradient descent minimization. We chose the optimal model in respect to CWAE (and respectively SWAE) cost, which consist of the sum of reconstruction error and the logarithm of CWAE (and respectively SWAE) distance.

We evaluated three types of batches, see Fig.~\ref{fig:cwae_celeba_comparison}, Tabs.~\ref{tab:comp_2} and~\ref{tab:comp_3}.


\begin{table*}[htb]
\normalsize
\caption{Comparison between proposed MoW mini-batch training procedure in two versions  ($n=64$, $k=1$ and $n=64$, $k=32$) with classical one ($n=k=64$). Grid search over learning rate parameter was used choosing optimal model in respect to CWAE cost (consisting of the sum of reconstruction error and the logarithm of  CWAE distance). See Fig.~\ref{fig:cwae_celeba_comparison} for FID scores.}
\begin{center}
{\small
\begin{tabular}[width=\textwidth]{lllrrrrr}  
\toprule
Data set & Method & Learing & CWAE & WAE & Rec. & CWAE & FID \\
 &  & rate & distance & distance & error & cost & score\\
\midrule             
MNIST    & CWAE $n=64$, $k=64$      & 0.01 & 0.024 & 0.008 & 5.617 & 1.903 & 48.201 \\
         & CWAE $n=64$, $k=32$  & 0.0075  & 0.024 & 0.008 & 5.324 & 1.604 &  43.796 \\
         & CWAE $n=64$, $k=1$  & 0.0025  &  0.032  & 0.025  & 4.989  & \bf 1.568 &  	36.960 \\                 
\midrule             
CIFAR10    & CWAE $n=64$, $k=64$ & 0.001 & 0.010 & 0.006 & 35.921 & 31.341 & 195.161 \\
         &  CWAE $n=64$, $k=32$ & 0.0025 &  0.009 & 0.005 & 37.285 & 32.651 & 159.279 \\
             & CWAE $n=64$, $k=1$  & 0.001 & 0.014 & 0.015 & 26.681 & \bf 22.435 & \bf 133.820 \\
\midrule             
CELEB~A & CWAE $n=64$, $k=64$ & 0.005 & 0.015 & 0.006 & 118.584 & 114.425 & 59.919 \\
      &  CWAE $n=64$, $k=32$ & 0.0075 & 0.014 & 0.004 & 117.466 & 113.204 & 58.577 \\
         &  CWAE $n=64$, $k=1$  & 0.00075 & 0.017 & 0.009 & 116.060  & \bf 112.025 & \bf 58.234\\
\bottomrule
\end{tabular}
}
\end{center}
\label{tab:comp_2}
\end{table*}

\begin{table}[htb]
\normalsize
\caption{Comparison between the MoW mini-batch training procedure for ($n=64$, $k=1$) and ($n=64$, $k=32$) with classical one ($n=k=64$) for SWAE model and various data sets. We used grid search over learning rate parameter and chose the optimal model in respect to SWAE cost, i.e. a sum of reconstruction error and the logarithm of SWAE distance.
See Fig.~\ref{fig:cwae_celeba_comparison} for FID scores.}
\begin{center}
{\small
\begin{tabular}[width=\textwidth]{lllrrrrr}  
\toprule
Data set & Method & Learing & CWAE & WAE & Rec. & WAE & FID \\
 &  & rate & distance & distance & error & cost & score\\
\midrule             
MNIST  & SWAE $n=64$, $k=64$ & 0.01 & 0.030 & 0.014 & 5.658 & 1.394 & 51.987  \\ 
         & SWAE $n=64$, $k=32$ & 0.001 & 0.031 & 0.016 & 7.291 & 3.164 & 79.401  \\
         & SWAE $n=64$, $k=1$ & 0.0025 & 0.095 & 0.156 & 4.871	& 3.018 & 43.398 \\
\midrule             
CIFAR10    & SWAE $n=64$, $k=64$ & 0.01 & 0.025 & 0.026 & 26.659 & 23.036 & 149.705 \\
         & SWAE $n=64$, $k=32$ & 0.001 & 0.049 & 0.058 & 25.295 & 22.457 & 228.323 \\
         & SWAE $n=64$, $k=1$ & 0.001 & 0.044 & 0.088 &  26.204 & 23.775 & 154.941 \\
\midrule             
CELEB~A & SWAE  $n=64$, $k=64$ & 0.001 & 0.026 & 0.020 & 121.953 & 118.072 & 73.150 \\
         & SWAE $n=64$, $k=32$ & 0.001 & 0.026 & 0.020 & 118.462 & 114.588 & 77.356 \\
         & SWAE $n=64$, $k=1$ & 0.00075 & 0.042 &0.066 & 115.683 & 112.970 & 79.518  \\
\bottomrule
\end{tabular}
}
\end{center}
\label{tab:comp_3}
\end{table}






\end{document}